%% file: 0_main.tex
\documentclass[5p,twocolumn]{elsarticle}
\usepackage{graphicx}
\usepackage{amsmath,amssymb} 
\usepackage{color}
\usepackage{amsmath,mathtools}
\usepackage[table]{xcolor}
\usepackage{multirow}
\usepackage{soul}
\usepackage{subcaption}
\usepackage{tabularx}
\usepackage[numbers]{natbib} 
\usepackage{epstopdf}
\usepackage{lineno}
\usepackage[hidelinks]{hyperref}
\usepackage{arydshln} 
\usepackage{comment}

\makeatletter
\def\ps@pprintTitle{%
   \let\@oddhead\@empty
   \let\@evenhead\@empty
   \let\@oddfoot\@empty
   \let\@evenfoot\@oddfoot
}
\makeatother

\biboptions{sort&compress}
\begin{document}  

\begin{frontmatter}
\title{OFF-ApexNet on Micro-expression Recognition System}

  \author[add1]{Sze-Teng Liong}
  \ead{stliong@fcu.edu.tw}
  \author[add2]{Y.S. Gan\corref{cor1}}
  \ead{ysgan@xmu.edu.my}
  \author[add3]{Wei-Chuen Yau}
  \author[add4]{Yen-Chang Huang}
  \author[add5]{Tan Lit Ken}
  
  \cortext[cor1]{Corresponding author}
  \address[add1]{Department of Electronic Engineering, Feng Chia University, Taichung 40724, Taiwan R.O.C.}
  \address[add2]{Department of Mathematics, Xiamen University Malaysia, Jalan Sunsuria, 43900 Sepang, Selangor, Malaysia} 
  \address[add3]{School of Information Science \& Engineering and Software, Xiamen University Malaysia, Jalan Sunsuria, 43900 Sepang, Selangor, Malaysia} 
  \address[add4]{School of Mathematics and Statistics, Xinyang Normal University, Henan, China} 
  \address[add5]{Malaysia-Japan International Institute of Technology (MJIIT), University Teknologi Malaysia Kuala Lumpur, Jalan Sultan Yahya Petra (Jalan Semarak), 54100 Kuala Lumpur, Malaysia} 

\begin{abstract}
When a person attempts to conceal an emotion, the genuine emotion is manifest as a micro-expression.
Exploration of automatic facial micro-expression recognition systems is relatively new in the computer vision domain.
This is due to the difficulty in implementing optimal feature extraction methods to cope with the subtlety and brief motion characteristics of the expression.
Most of the existing approaches extract the subtle facial movements based on hand-crafted features.
In this paper, we address the micro-expression recognition task with a convolutional neural network (CNN) architecture, which well integrates the features extracted from each video. 
A new feature descriptor, Optical Flow Features from Apex frame Network (OFF-ApexNet) is introduced. This feature descriptor combines the optical flow guided context with the CNN.
Firstly, we obtain the location of the apex frame from each video sequence as it portrays the highest intensity of facial motion among all frames.
Then, the optical flow information are attained from the apex frame and a reference frame (i.e., onset frame).
Finally, the optical flow features are fed into a pre-designed CNN model for further feature enhancement as well as to carry out the expression classification.
To evaluate the effectiveness of OFF-ApexNet, comprehensive evaluations are conducted on three public spontaneous micro-expression datasets (i.e., SMIC, CASME II and SAMM).
The promising recognition result suggests that the proposed method can optimally describe the significant micro-expression details. 
In particular, we report that, in a multi-database with leave-one-subject-out cross-validation experimental protocol, the recognition performance reaches 74.60\% of recognition accuracy and F-measure of 71.04\%. We also note that this is the first work that performs cross-dataset validation on three databases in this domain.

\end{abstract}


\end{frontmatter}


\input{1_intro}

\input{2_related}

\input{3_FE}
\input{4_algo}

\input{5_experiment}

\input{6_results}
\input{7_conclusion}

\section*{References}
\bibliography{mybibfile}

\end{document}

%% file: 1_intro.tex
\section{Introduction} 
Facial expression is one of the popular nonverbal communication types that plays an important role to reflect one's emotional state.
Different combinations of facial muscular movement eventually represent specific type of emotions. 
According to the psychologists, people portray some particular emotions on the face in the same way, regardless the race or culture \citep{ekman1971constants}.
Furthermore, it was verified by ~\cite{matsumoto2009spontaneous} that there is no difference between the sighted and blind individuals on the configuration of the facial muscle movements to response to the emotional stimuli.
In other words, facial expressions are universal.
They can be commonly classified into six emotion classes: happiness, sadness, fear, anger, disgust and surprise. 

Generally, facial expression is categorized into two types, namely, macro-expression and micro-expression.
The formal expression typically lasts between three
quarters of a second to two seconds, and the muscle movements are possibly
occurred simultaneously at multiple parts on the face.
Therefore, macro-expressions are readily perceived by humans in real time conversations.
Over the past few decades, the research in automated macro-expression recognition analysis has been an active topics.
To date, plenty of the recognition systems developed achieved more than 95\% of expression classification accuracy \citep{lopes2017facial,wang2016facial} and some of them even reached almost 100\% perfect recognition performance \citep{kharat2009emotion, ali2015facial, rivera2013local}.
However, it should be noted that macro-expression does not accurately implies one's emotion state as it can be easily faked. 
Hence, it is worth to investigate to deeper emotion states from the muscular movements.

Among several types of nonverbal communications, micro-expressions are discovered to be more likely to reveal one's true emotions. 
Micro-expressions often sustain within one-twenty-fifth to one-fifth of a second \citep{ekman1971constants} and they may only present in a few small regions on the face.
Besides, they are stimulated involuntary which means that people cannot control their appearance.
This allows the competent in exposing one's concealed genuine perceptions without deliberately control.
Owing to its characteristic of potentially exposing a person's true emotions, it can be deployed in several applications such as national security, police interrogation, business negotiation, social interaction and clinical practice \citep{seidenstat2009protecting, o2009police, matsumoto2011evidence, turner1997evolution, frank2009see}.

Micro-expressions were first discovered by ~\cite{haggard1966micromomentary} about fifty years ago, when analyzing on a couples of psychotherapeutic interviews films.
At that time, they referred to the expression as ``micro momentary expression (MME)" and its appearance is the result of a repression feeling.
A few years later, ~\cite{ekman1969nonverbal} did a groundbreaking discovery when watching on a slow-motion interview film of a depressed patient who was requesting for a weekend pass from the psychiatric hospital to go home.
Through a carefully frame-by-frame observation on the video, Ekman and Friesen noticed the appearance of strong negative intense micro-expressions that the patient was trying to hide.
However, the emotions were quickly covered up with another expressions (i.e., smile). 
In fact, the patient was planned to commit suicide without the supervision.
Since then, analysis in micro-expression is gaining more attention in both the psychological and computer vision fields.

Thus far, the identification and annotation of micro-expressions are done manually by psychologists or trained experts.
This may lead to reliability inconsistency as the labeling of the expression is solely dependent on the personal judgment.
In addition, it is time and effort consuming as the annotators are required to inspect the tiny facial muscle changes in each frame transition.
Therefore, it is essential to implement reliable computer-based micro-expression detection and classification systems to obtain trustable, accurate and precise ground-truths (i.e., emotion state, action unit, onset, apex and offset indices) of each video. 

\begin{figure}[tb]
\centering
\includegraphics[width=1\linewidth]{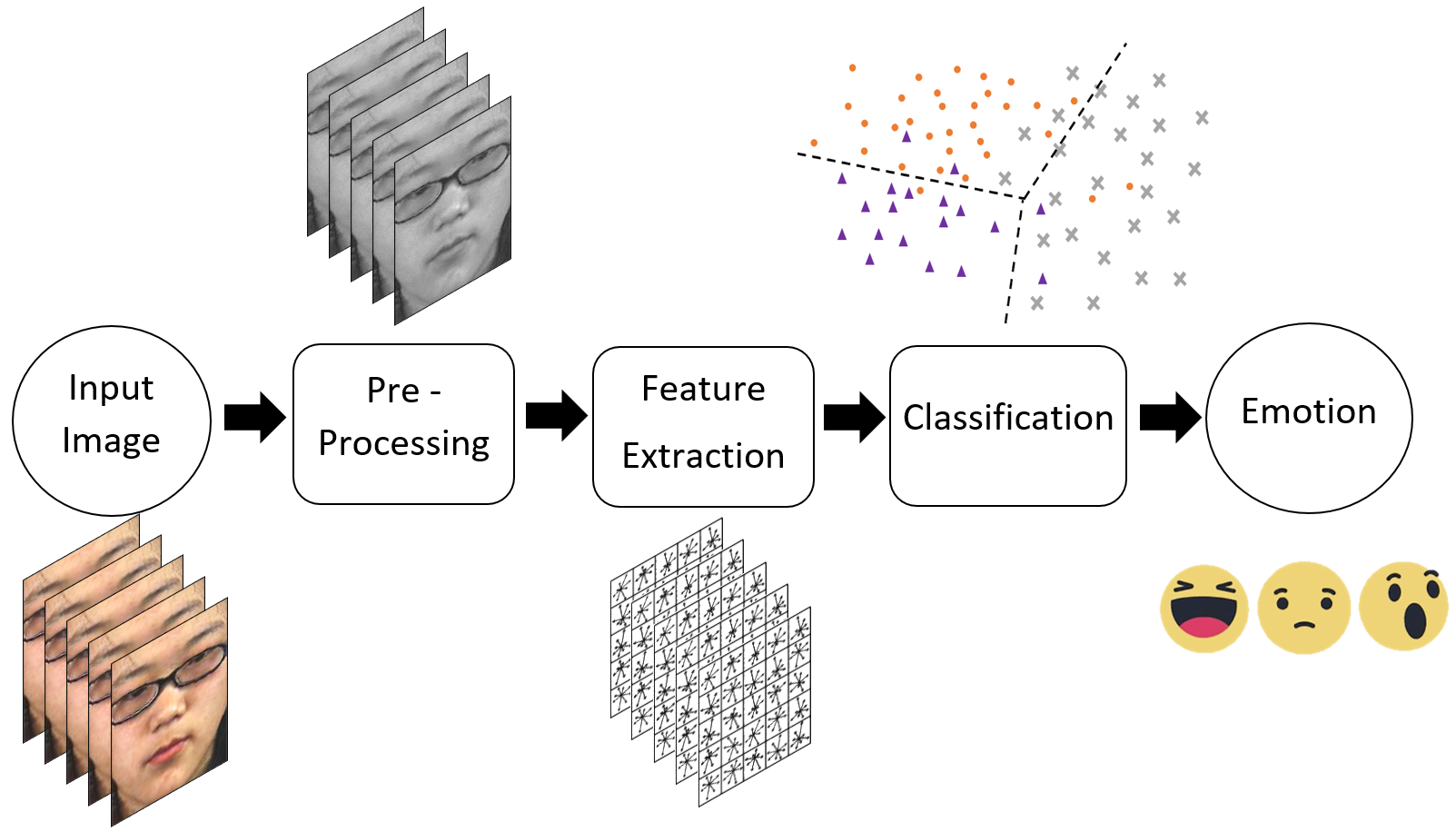}
\caption{Block diagram of a typical facial micro-expression recognition system}
\label{fig:basicStep}
\end{figure}

In general, a micro-expression recognition system involves three basic steps, include: 
(1) Image preprocessing - enhancement of image by preserving the significant features;
(2) Feature extraction - identification of the important features from the image;
(3) Expression classification - recognition of the emotion based on the features extracted. 
Figure~\ref{fig:basicStep} illustrates the basic flowchart of the recognition process.
Each step plays a vital role to obtain a promising recognition performance and they are all equivalently important because each of them is targeting unique strategies to address the desired features in different perspective.
In the recent years, the automated micro-expression systems developed in the literature are increasing gradually.
This might due to the lack of suitable databases for data training and testing purposes, and hence hindering further analysis study especially in performance assessment and investigation.
To date, there are three spontaneous publicly-available micro-expressions databases (i.e., CASME II \citep{casme2}, SMIC \citep{smic} and SAMM \citep{samm}) that contain sufficiently large number of video samples for experimental evaluation.

Recent works ~\citep{patel2016selective, takalkar2017image, peng2017dual} have shown the feasibility of adopting deep learning (e.g., convolutional neural network (CNN)) in micro-expression recognition systems. However, the recognition accuracy of previous works are still unsatisfactory.

To the best of our knowledge, there has not been any attempt that performs cross database evaluation for micro-expression recognition task using CNN mechanism.
In this paper, a novel and robust feature extraction approach that can effectively represent the subtle facial muscle contractions for micro-expression recognition system is presented.
Concretely, the contributions of this paper are listed as follows:
\begin{enumerate}
   \item Adoption of only two frames (i.e., onset and apex) from each video to better represent the significant expression details and applying optical flow guided techniques to encode the motion flow features.
   \item Proposal of a novel feature extractor that incorporates both the handcrafted (i.e., optical flow) and data-driven (i.e., CNN) features.
   \item Implementation of a novel CNN architecture that is capable to highlight valuable input features and improve the emotion state prediction.
   \item Comprehensive evaluation of the proposed approach on three recent spontaneous micro-expression databases is performed to validate its consistency and effectiveness.
\end{enumerate}

The remainder of the paper is organized as follows.
Section~\ref{sec:related} discusses related works on the state-of-the-art apex frame spotting and feature extraction techniques.
The proposal of the recognition system framework, theoretical derivations and the effective use of CNN are elaborated in Section~\ref{sec:algorithm}. 
Overview of the databases used and the experimental settings are described in Section~\ref{sec:experiment}.
Followed by Section~\ref{sec:results} that reports the recognition performance, with discussion and analysis.
Finally, conclusions are drawn in Section~\ref{sec:conclusion}.

%% file: 2_related.tex
\section{Related Work}
\label{sec:related}

In the literature, most of the automated micro-expression studies focused on the first and second stages of the recognition system, i.e., image preprocessing and feature extraction.
Some promising preprocessing techniques and feature extractors exploited in micro-expressions analysis systems will be discussed and elaborated in the following subsections.

\subsection{Image Preprocessing}
The two properties of the micro-expressions are low intensity and often occur in specific facial regions.
Therefore, some of the previous works aim to emphasize the facial muscle movements in some particular areas, instead of extracting the features from the entire face.
By focusing to extract features from several small facial regions can omit the noticeable background noises captured by the camera (which are probably due to the flickering lights).
In addition, considering the regions of interest (RoIs) is able to accelerate the feature extraction and classification processes as irrelevant data are eliminated.
For instance, ~\cite{wang2014micro} encode the expression features from 16 RoIs based on the Facial Action Coding System (FACS) \citep{ekman1978facial} which indicate the relation between the facial muscle changes and the emotion state.
However, the shapes and sizes of the 16 RoIs are not flexible as they are heavily rely on the feature coordinates detected by the landmark detector.
On the other hand, ~\cite{liong2018hybrid} proposed to reduce number of RoIs to three regions (i.e., ``left eye + left eyebrow", ``right eye + right eyebrow" and ``mouth").
The selection of these three areas are identified according to the occurrence frequency of the muscle movements in the videos provided by CASME II and SMIC databases.
Although the size and location of the 3 RoIs are not fixed, they are merely dependent on the position of the landmark coordinates. 
Unfortunately, the landmark-based approach might not be sufficiently accurate and the 3 regions selected are not always the optimal areas that can capture the perfect expression information.
In addition, it is pointed by~\cite{xu2017microexpression}, that a fine-scale alignment is essential to be performed as the preprocessing step.
This is because the subtle misalignment resulted from the conventional facial registration and alignment tools could cause degradation in the recognition performance.
 
Moreover, there are some works that minimize the information redundancy in micro-expressions by emphasizing only a portion of all frames of each video.
For example, ~\cite{le2017sparsity} select several important frames for extraction.
This is intuitive as the images are captured using high frame rate cameras, there will be similar facial motion patterns appearing in consecutive frames.
Therefore, they intent to identify and remove unfavorable redundant frames as the preprocessing step.
Besides, this could boost the discrimination power of the feature vectors.
On a similar note, another recent method proposed by ~\cite{he2017multi} also describes the expression details from a reduced set of frames.
Concretely, Temporal Interpolation Model (TIM) ~\citep{zhou2012image} is applied to normalize all the videos in SMIC dataset to 20 frames and CASME II to 30 frames.
It should be noted that the average frame length for SMIC and CASME II are 33 and 67, respectively.
Although shorten the video length improves the efficiency and accuracy performance, an arbitrary decision has to be made about what frame length should be used.

Another remarkable preprocessing technique proposed in~\cite{liong2018less,liong2017micro} well-represent the entire video by utilizing only the apex frame (and onset frame as reference frame).
To be concise, there are generally three temporal segments in each micro-expression videos (i.e., onset, apex and offset).
The onset is the instant that the facial muscles begins to contract and grow stronger.
The apex frame indicate the most expressive facial action when it reaches the peak.
The offset is the moment where the muscles are relaxing and the face returns to its neutral appearance.
From the results reported in~\cite{liong2018less,liong2017micro}, it supports that, encoding the features from apex frame provides more valuable expression details than a series of frames.
Furthermore, the apex-based approach is employed in the other work of~\cite{liong2016automatic}, where they tested on other micro-expression databases comprising only of raw long videos and promising performance results are obtained.

\subsection{Feature Extraction}
The primitive feature extraction method that evaluated on spontaneous micro-expression databases (i.e., CASME II and SMIC) is known as Local Binary Pattern on Three Orthogonal Planes (LBP-TOP) \citep{zhao2007dynamic}.
LBP-TOP was eventually designed to describe dynamic texture patterns.
In brief, it is capable to capture the local spatio-temporal motion information (i.e., pixel, region and volume levels).
Furthermore, it is robust against geometric variations caused by scaling, rotation, or translation.
With great discriminative feature representation as well as its computational simplicity, LBP-TOP has been comprehensively studied and modified to accommodate in different applications.
As a result, several LBP variants are proposed and some of them are  examined in micro-expression analysis, such as Local Binary Patterns with Six Intersection Points (LBP-SIP) \citep{wang2014lbp}, Spatiotemporal Local Binary Pattern with Integral Projection (STLBP-IP) \citep{huang2015facial}, Completed Local Quantization Pattern (STCLQP) \citep{huang2016spontaneous}.

Apart from LBPTOP, optical flow \citep{gibson1950perception} is one of the most popular feature extractors, as it has been very successful in a variety of computer vision tasks, such as action recognition \citep{chaudhry2009histograms}, face tracking \citep{decarlo2000optical},  medical image reconstruction \citep{weng1997three}.
Succinctly, optical flow measures the apparent motion of the brightness patterns in a sequence of images in terms of velocity vector field.
Owing to its robust feature representations with data from multiple domains, a number of researchers unleashed the potential of optical flow in micro-expression recognition systems.
For instance, ~\cite{liu2016main} proposed to construct a RoI-based feature vector using optical flow to describe the local motion information and the spatial location.
Thus, aside having compact feature representation (i.e., feature dimension of 72 per video), it is robust to translation, rotation and illumination changes.
As an extension of optical flow, ~\cite{shreve2009towards} derived a higher order accurate differential approximation, namely optical strain. 
Optical strain leads to better performance in determining the motion changes compared to optical flow, as it is capable to preserve relatively meaningful facial muscle movements \citep{liong2016spontaneous, liong2014subtle}.

Deep learning has emerged as a family of machine learning technique that operates such that the important information are iteratively extracting from data and transforming them into the final output features.
Deep learning has significant impacts on a variety of application domains as it yields numerous state-of-the-art results, such as speech recognition \citep{amodei2016deep}, face recognition \citep{sun2015deepid3} and scene recognition \citep{zhou2014learning}.
However, deep learning has yet to have a widespread impact on micro-expression studies.
Particularly, the first work that adopts Convolutional Neural Network (CNN) is established by ~\cite{patel2016selective} to evaluate the proposed algorithm in CASME II and SMIC databases with Leave-One-Subject-Out Cross Validation (LOSOCV) protocol during the data training and testing stages.
However, the accuracy results obtained by their work do not outperform the conventional methods as the model is possibly being overfitted.
Besides, ~\cite{takalkar2017image} intent to increase the number of samples to the double of each dataset using data augmentation.
They partitioned all the images into three sets, namely training, testing and validation, which consist the portion of 80\%, 1\% and 1\%, respectively. 
On the other hand, a recent work by~\cite{peng2017dual} directly feeds the CNN model with high level features (i.e., optical flow). 


%% file: 3_FE.tex

%% file: 4_algo.tex
\section{Proposed Algorithm}
\label{sec:algorithm}

The impressive recognition performance presented in the earlier work by ~\cite{liong2018less} has brought the significance of the apex frame into a sharp focus, especially in the feature extraction stage.
With rich motion patterns obtained from the apex frame (with onset as the reference frame), it is possible to select the features with minimal redundancy.
As a result, the facial regions containing relevant details of the expression can be easily noticed and encoded. 

\begin{figure*}[tb]
\centering
\includegraphics[width=1\linewidth]{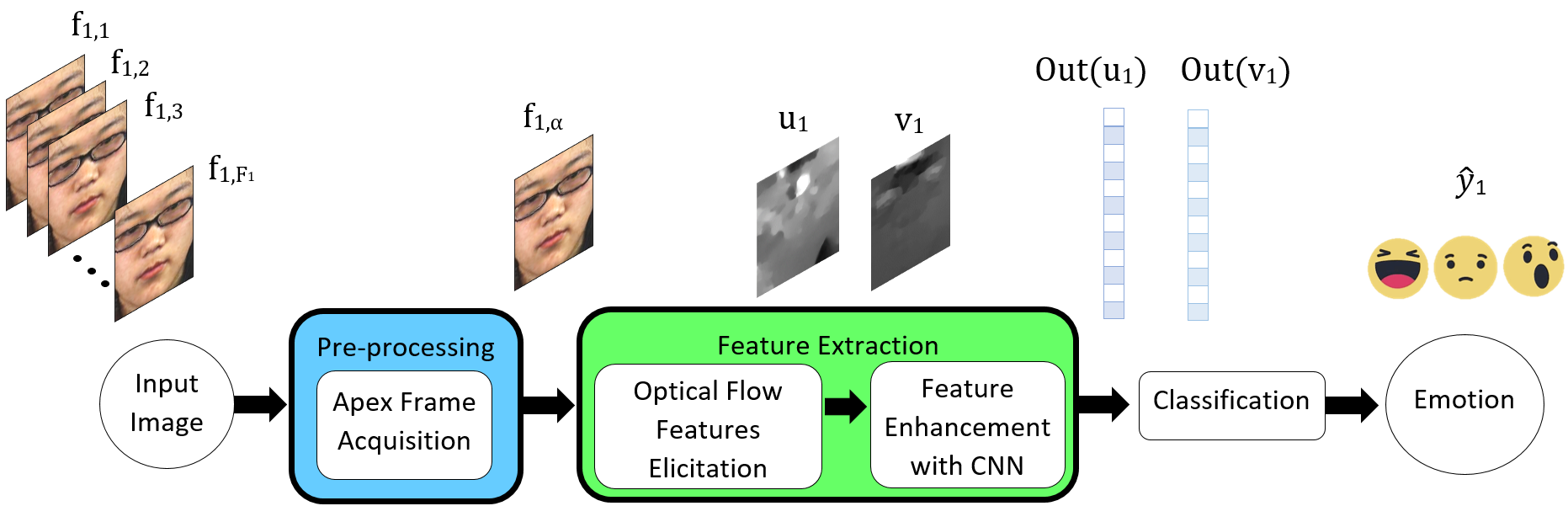}
\caption{Overview of the proposed micro-expression recognition system. It consists of three main steps, namely apex frame acquisition, optical flow features elicitation and feature enhancement with CNN.}
\label{fig:proposedFlow}
\end{figure*}

The proposed method is targeted to emphasize on the preprocessing and feature extraction stages.
In brief, it incorporates the following three steps:
\begin{enumerate}
\item Apex frame acquisition: to spot the apex frame location from each video.
\item Optical flow features elicitation: to estimate the horizontal and vertical optical flow from the apex and onset frames.
\item Feature enhancement with CNN: to enrich the optical flow features that can automatically identify and learn relevant spatio-temporal context information in a hierarchical way.
\end{enumerate}
A conceptual framework in this paper is illustrated in Figure~\ref{fig:proposedFlow}.
The detailed procedures for each step are described in the following subsections.

\subsection{Apex Frame Acquisition}
\label{subsec:apex}
There are three micro-expression databases exploited in the experiment, namely, CASME II \citep{casme2}, SMIC \citep{smic} and SAMM \citep{samm}. 
The location of the ground-truth apex frame has been provided in CASME II and SAMM, which are annotated by at least 2 trained experts.
Since the apex frame index in SMIC is absence, an automatic apex spotting system has to be applied to approximate the location of apex frame.
It has been demonstrated that the apex spotting mechanism, D\&C-RoIs \citep{liong2015automatic}, is capable to exhibit reasonable good recognition performance \citep{liong2018less,liong2016automatic}.
Succinctly, the D\&C-RoIs method first computes the LBP features from three facial sub-regions (i.e., ``left eye+eyebrow", ``right eye+eyebrow" and ``mouth") of each image. 
Then, a correlation coefficient principle is employed to acquire the changes in difference of the LBP features between the onset frame to the rest of the frames.
Finally, a Divide \& Conquer strategy is utilized on the rate of the feature difference to search for the apex frame, whereby it indicates the frame index of the local maximum.

For clarity, the notations used in this paper are defined and explained in the following sections.
A micro-expression video clip is expressed as:
\begin{equation}
S = \left[ s_1, s_2, ... , s_n\right],
\end{equation}

\noindent where $n$ is the number of video clips.  
The $i$-th of the sample video clip is molded to:
\begin{equation}
s_{i} = \{f_{i,j} | i=1,\dots,n; j=1,\dots ,F_{i}\},
\end{equation}

\noindent where $F_i$ is the total number of image frames in the $i$-th sequence.
There will be one apex frame in each video sequence and it can be located at any frame index between the onset (first frame) and offset (last frame). 
The onset, apex and offset frames are denoted as $f_{i,1}$, $f_{i,\alpha}$ and $f_{i,F_i}$, respectively.  
The apex frame can be denoted as:
\begin{equation}
f_{i,\alpha} \in f_{i,1},\dots ,f_{i,F{i}}
\end{equation}

Thus, $f_{i,\alpha}$ is predicted after adopting the D\&C-RoIs approach.

\subsection{Optical Flow Features Elicitation}
\label{subsec:elicitation}
In this process, a higher level with reduced dimension features are produced in this stage.
Consequently, the optical flow features are obtained prior to passing the raw onset and apex images to the CNN architecture.
Optical flow is able to indicate the apparent facial motion changes between frames. 
It is an approximation of the image patterns based on the local derivatives between two images.
Specifically, it aims to generate a two-dimensional vector field, i.e., motion field, that represents the velocities and directions of each pixel.
In order to attain the dynamical movement of the desired optimal expression (i.e., $p_{i,\alpha}$), the intensity difference between the onset (i.e., $f_{i,1}$) and apex (i.e., $f_{i,\alpha}$) is estimated.

To estimate the optical flow, it is generally assumed that: 
\begin{itemize}
\item The apparent brightness of the moving objects remains unchanged between the source and target frames.
Thus the noises generated by a large variety of imaging variables such as the shadows, highlights, illumination and surface translucency phenomena are entirely neglected.
\item The movement between two consecutive frames are small as the motion changes gradually over time.
\item Image flow field is continuous and differentiable in both the space and time domains.
\item The scene is static, the objects in the scene are rigid, and the changes of the objects' shape are ignored.
\end{itemize}


Suppose that the intensity of the reference frame that locates at $t$-th of a video sequence is defined as $I_t(x,y)$.
The intensity of the next consecutive
frame, $(t+1)$-th is denoted as $I_{t+1}(x+\delta x, y + \delta y)$.  
According to the brightness constancy constraint, the intensity of the two adjacent frames is achieved as:
\begin{equation}
\label{eq:It}
I_t(x,y) = I_{t+1}(x + \delta x, y + \delta y),
\end{equation}

\noindent where $\delta x = u^t \delta t$ and $\delta y = v^t \delta t$.
Explicitly, $u^t(x,y)$ and $v^t(x,y)$ refer to the horizontal and vertical of the optical flow field, respectively.    
By adopting Taylor series expansion on (\ref{eq:It}), it becomes an expanded form: 

\begin{equation}
\label{eq:It2}
I_{t+1}(x + \delta x, y + \delta y) \approx I_t(x,y) + \delta{x} \frac{\partial{I}}{\partial{x}}+\delta{y} \frac{\partial{I}}{\partial{y}} + \delta {t} \frac{\partial{I}}{\partial{t}} 
\end{equation}

We then combines (\ref{eq:It}) and (\ref{eq:It2}), the optical flow equation can be succinctly formulated as follows:

\begin{equation}
\begin{split}
I_t(x,y) &=  I_t(x,y) + \delta{x} \frac{\partial{I}}{\partial{x}}+\delta{y} \frac{\partial{I}}{\partial{y}} + \delta {t} \frac{\partial{I}}{\partial{t}}, \\
0 &=  \delta{x} \frac{\partial{I}}{\partial{x}}+\delta{y} \frac{\partial{I}}{\partial{y}} + \delta {t} \frac{\partial{I}}{\partial{t}}
\end{split}
\end{equation}

\noindent By dividing both sides of the equations by $\delta t$:
\begin{equation}
\begin{split}
0 &=  \frac{\delta{x}}{\delta{t} } \frac{\partial{I}}{\partial{x}}+\frac{\delta{y}}{\delta{t} }\frac{\partial{I}}{\partial{y}} + \frac{\delta{t}}{\delta{t} }\frac{\partial{I}}{\partial{t}},\\
0 &= u^t(x,y)\frac{\partial I}{\partial x} + v^t(x,y)\frac{\partial I}{\partial y} + \frac{\partial I}{\partial t}
\end{split}
\end{equation}

For a sufficiently small interval time between the onset and apex frames (i.e., less than 0.2 seconds), it is assumed that the brightness of the surface patches remains constant.
Hence, the optimal expression flow feature $p_{i,\alpha}$, can be obtained 
as: 
\begin{equation}
I_{t = 1} (x,y) = I_{t + \alpha}(x + u^t(x,y)\delta t, y+ v^t(x,y) \delta t)
\end{equation}

Finally, the optical flow map that computed from the two frames (i.e., onset and apex) is formed to represent the entire video: 
\begin{equation}
O_i = \{(u(x,y), v(x,y)) | x = 1, 2, ... , X, y = 1, ... , Y\},
\end{equation}
\noindent where X and Y denote the width and height of the images, $f_{i,j}$, respectively.

In short, each video sequence, $s_i$ is summarized into the following two optical flow derived representations:
\begin{enumerate}
\item  $u(x,y)$  - Horizontal component of the optical flow field $O_i$
\item $v(x,y)$ - Vertical component of the optical flow field $O_i$

\end{enumerate}

The optical flow technique utilized in the experiments later is TV-L1 \citep{zach2007duality} method.
This is because it is better in preserving the flow discontinuities and is more robust compared to the classical optical flow method (i.e., Black and Anandan \citep{black1996robust}) \citep{shreve2009towards}.

\subsection{Feature Enhancement with Convolutional Neural Network}
\label{subsec:enhancement}
The optical flow features contain the spatio-temporal expression details. They are then fed into a CNN architecture which is expected to further improve the feature information by reconstructing and refining the selection of more significant motion details.
CNN is one of the deep artificial neural networks that has been widely used in analyzing visual imagery \citep{tu2018multi, bai2018sequence, ullah2018action}. 
It consists of several layers, such as the input layer, convolutional layer, pooling layer, fully connected layer and output layer. 
CNN has also been recently exploited in micro-expression recognition mechanisms.
For example, ~\cite{peng2017dual} designed a 3D-CNN architecture to effectively learn the high-level features (i.e., optical-flow data).
However, in contrast to~\cite{peng2017dual}, the optical flow representations obtained from the previous stage (i.e., Section~\ref{subsec:elicitation}) are having two dimensional maps (i.e., $X\times Y$).
Therefore, a new 2D-CNN architecture is proposed to perform the feature learning task.

\begin{figure*}[tb]
\centering
\includegraphics[width=1\linewidth]{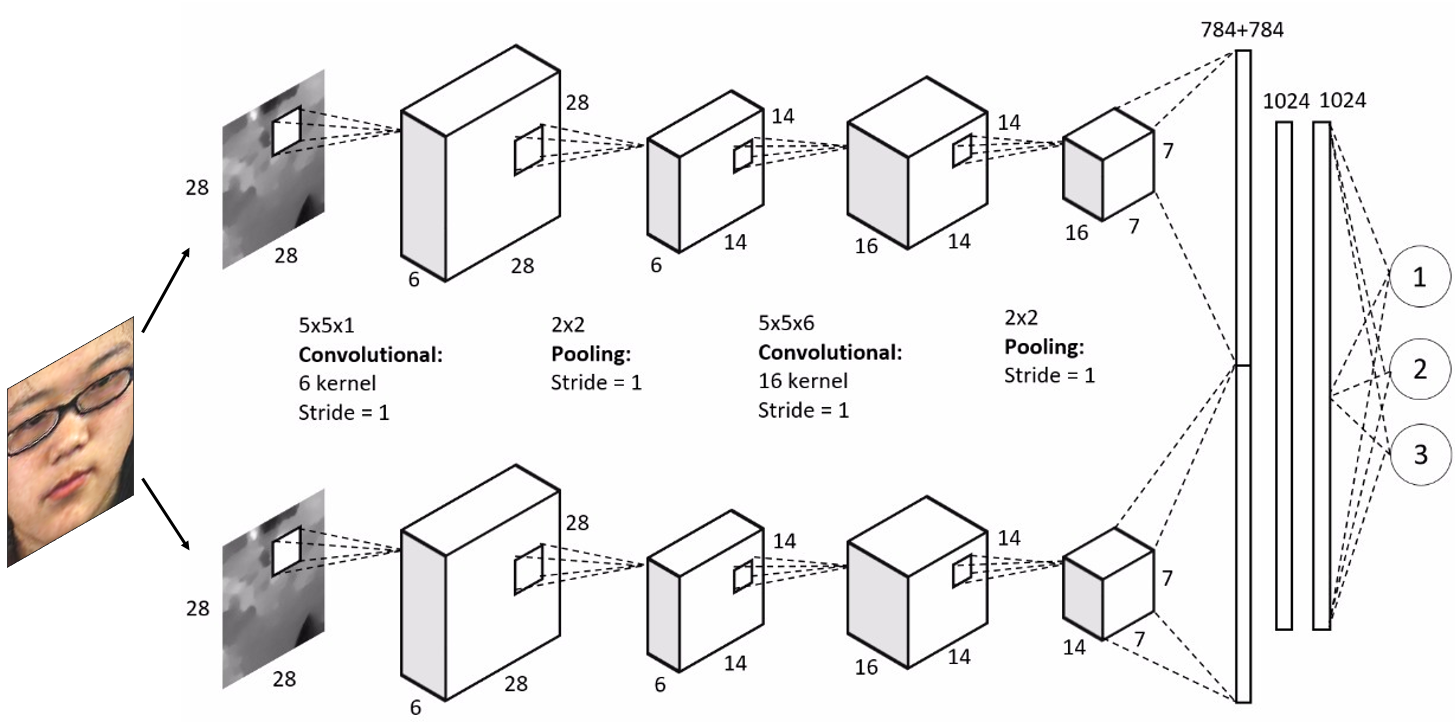}
\caption{Framework of the proposed OFF-ApexNet architecture.
The input data is the horizontal and vertical optical flow images.
They are then processed by two convolutional layers and two pooling layers, followed by two fully connected layers.}
\label{fig:proposedCNN}
\end{figure*}

Figure~\ref{fig:proposedCNN} illustrates the conceptual visualization of our proposed OFF-ApexNet (Optical Flow Features from Apex frame Network) 
architecture. The horizontal and vertical components of the optical flow are used as the input data of the CNN.
Two independently trained CNN models (i.e., to train \textit{u} and \textit{v} separately) will be merged to form a resultant feature vector at the fully connected layers. 
The basic overview of the duty of each layer is described and explained as follows.

First, for the input layer, all the input data are normalized to a fix size (i.e., $\aleph\times \aleph$), whereby the input data in this case is the optical flow based components, such that:
\begin{equation}
u= \frac{\delta x(t)}{\delta t},
\end{equation}

\noindent and
\begin{equation}
v = \frac{\delta {y (t)}}{\delta y}, \end{equation}

\noindent where $u$ and $v$ refers to the horizontal and vertical components of optical flow, respectively.
The normalized data is then multiplied with a convolution kernel to form a feature map in the following convolutional layer.
Concretely, each $e_{ij}$ pixel in the feature map is calculated by:
\begin{equation}
\begin{split}
e^{l}_{ij} &= \{f^l(x^{l}_{ij} + b^{l})| i = 1, 2, ... , \aleph, j = 1, ... , \aleph\},\\
 \text{where } x^{(l)}_{ij} &= \Sigma^{m-1}_{a=0}\Sigma_{b=0}^{m-1}w^{(l)}_{ab} y^{l-1}_{(i+a)(j+b)},
\end{split}
\end{equation}

\noindent $x^{(l)}_{ij}$ is the pixel value vector of the set of units in the small neighborhood corresponding to $e_{ij}$ pixel at layer $l$, whereas $f^l$ denotes the ReLu activation function at layer $l$.
$w$ and $b$ are the coefficient vector and bias respectively, determined by the feature map.   
Thus for an input $x$, the ReLu function can be indicated as:
\begin{equation}
f(x) = max(0, x)
\end{equation}

The input optical flow features (i.e., $u$ and $v$) are now transformed into feature maps (i.e., $e$) representation. 
The size of generated feature map is rely on the number of convolution kernels.
Conventional kernel sizes chosen in the past research are $3\times 3$, $5\times 5$ and $7\times 7$.

The subsequent layer is the pooling layer.
It is used as a subsampling operator to progressively reduce the spatial size of the feature map representation.
As a result, it can effectively minimize the computational complexity of the CNN architecture.  
The $k$-$th$ unit in the feature map in the pooling layer can be achieved by: 
\begin{equation}
Pool_k = f(down(C)*W + b),
\end{equation}

\noindent where $W$ and $b$ are the coefficient and bias, respectively.
$down(\cdot)$ is a subsampling function, which can be expressed as:
\begin{equation}
down(C) = max\{C_{s,l} | s\in Z^+, l \in Z^+ \le m\},
\end{equation}

\noindent where $C_{s,l}$ refers to the pixel value of C in the feature map $e$.
$m$ denotes the sampling size.  

It is observed that each layer (i.e., convolutional layers and pooling layers) in the CNN architecture deliberately learn and convert the optical flow features to higher level features in other subsequence layers. 
After passing through all the convolution network layers (which may consists of several convolution layers and pooling layers), the final feature representation (denoted as $Out(\tau)$) comprises significant expression information, where $\tau$ is the optical flow based features of input images (i.e., $u$ and $v$).  

Since the total number of videos used in the experiments is relatively few (i.e., 441 from three datasets), the proposed CNN architecture is composed of only four layers (i.e., two convolution layers and two pooling layers).
These layers are responsible to generate meaningful features from the input data, where the final output $Out(\tau)$ can be concisely expressed as follows:
\begin{equation}
\begin{split}
Out(u) =& f^4(down(f^3( ( f^2(down(f^1(u \ast W^1 + b^1)) \\ &\ast W^2 + b^2))\ast W^3+b^3)) \ast W^4 + b^4)
\end{split}
\end{equation}

\noindent and 
\begin{equation}
\begin{split}
Out(v) =& f^4(down(f^3( ( f^2(down(f^1(v \ast W^1 + b^1)) \\ &\ast W^2 + b^2))\ast W^3+b^3)) \ast W^4 + b^4)
\end{split}
\end{equation}

The high-level reasoning features (i.e., $Out_u$ and $Out_v$) derived from the input data are then flattened and merged tbefore passing to the following fully connected layer.
In general, the fully connected layers transforms the features to the a set of desired number of classes from the analysis of frequencies based on the importance of features.
There are three emotion classes in the experiments, namely positive, negative, and surprise.
Note that similar to the convolutional layer, a ReLu activation function is applied to to all of the output after the fully connected layer.

Next, the transformed features from the fully connected layer is passed into the output layer.
The amount of neurons in the output layer is associated with the number of classes to be classified, which is three in this case.
The output probabilities of each class are computed using an activation function, which will result to a sum of one.
However, practically the output given by the former layers do not guarantee that the total sum of the probabilities over all classes equals to one.
To resolve this issue, a softmax regression is utilized as the activation function.  
Specifically, the probability of classifying into class $c$ is given by:
\begin{equation}
\hat{y} = p\left(y = c | x_j \right) = \frac{e^{x_j}}{\Sigma_{n = 1}^{N}e^{x_n}},  1\le c \le C,
\end{equation}
where $y$ is the ground-truth value of input $x_j$, C is the number of the classes. The loss function can be defined as follows:
\begin{equation} \label{eq1}
L(y, \hat{y}) = - \Sigma_{i=1}^N l(y_i) \log(\hat{y_i}),
\end{equation}

\noindent where $l\{\cdot\}$ is eigenfunction.
When $l\{\cdot\}$ is true, the loss function will return a number of one.   
The gradient of error can be calculated using (\ref{eq1}).   
Then sum of errors from multiple inputs is anticipated to be minimized by updating the weights of networks using a stochastic Adam gradient descent.
This particular type of gradient descent is known as an optimization algorithm.
It aims to search for the weights and coefficient in the neural network by performing backpropagation, so that the actual output to be closer the target output.
Thereby, decreases the error of each output neuron and the network as a whole. 

%% file: 5_experiment.tex
\section{Experiment}
\label{sec:experiment}
\subsection{Database}
There are a total of three micro-expression databases involve in the experiment, namely SMIC \citep{smic}, CASME II \citep{casme2} and SAMM \citep{samm}. 
This is to avoid the issue of overfitting, which will happen when the gap between training and testing errors is large.
Since the number of video of each single database is considered small (i.e., $\approx$ 150), it will fit the training dataset very well but underperform on new datasets.
Besides, more training data can improve the data generalization capability.
As such, by considering all the three datasets as a whole, it could lead to constructing a good predictive model.
Thus, better in recognizing the new (i.e., unseen) faces with different imaging conditions and environments.

Note that the databases are being preprocessed prior to releasing to the recorded videos to the public.
For instance, facial alignment is carried out in order to standardize all the faces into a uniform size and shape. 
Besides, it is also to ensure that the data extracted later are capable of integration. 
Succinctly, face alignment is a process of detecting the transforming a set of landmark coordinates to map the face to the model face.
Specifically, both the SMIC \citep{smic} and CASME II \citep{casme2} utilized Active Shape Model (ASM) \citep{van2002active} to allocate the 68 facial landmark points then Local Weighted Mean (LWM) \citep{goshtasby1988image} is employed to transform the faces based on the model face.
For SAMM, the faces are first registered with a Face++ automatic facial point detector \citep{Face}, then dlib \citep{king2009dlib} is adopted as the face alignment tool.

An overview of the micro-expression datasets information that used in the experiment is shown in Table~\ref{table:database}.
More details are elaborated as follows.

\begin{table}[tb]
\begin{center}
\caption{Detailed information of the SMIC, CASME II and SAMM databases used in the experiment}
\label{table:database}
\begin{tabular}{llccc}
\noalign{\smallskip}

\cline{3-5}
\noalign{\smallskip}
 \multicolumn{2}{l} {} & SMIC & CASME II & SAMM  \\
\hline
\noalign{\smallskip}
\multicolumn{2}{l}{Participants}
&	16 & 24 & 28\\
\hline
\noalign{\smallskip}
\multicolumn{2}{l}{Frame rate (\textit{fps})}
&	100 &  200 & 200  \\
\hline
\noalign{\smallskip}
\multicolumn{2}{l}{\multirow{2}{*}{\begin{tabular}[c]{@{}l@{}}Cropped resolution\\  (pixels)\end{tabular}}}
&
\multicolumn{3}{c}{\multirow{2}{*}{\begin{tabular}[c]{@{}l@{}} $170 \times 140$\end{tabular}}} \\

\multicolumn{2}{l}{}                                   &  \multicolumn{3}{l}{}  \\

\hline      
\noalign{\smallskip}
\multicolumn{2}{l}{Avg. frame number}
&	34 & 68 & 74 \\
\hline
\noalign{\smallskip}
\multicolumn{2}{l}{Avg. video duration (\textit{s})}
&	0.34 & 0.34 & 0.37 \\
\hline
\noalign{\smallskip}
\multirow{4}{*}{Expression} 
& 	Negative	&	70	&	88 &	91\\
& 	Positive	&	51	&	32 &	26\\
& 	Surprise	&	43	&	25 &	15\\
& 	Total	&	164	&	145&	132\\
\hline
\noalign{\smallskip}
\multirow{3}{*}{\begin{tabular}[c]{@{}l@{}}Ground-truth\\ (index)\end{tabular}}
& 	Onset	&	Yes	&	Yes &	Yes\\
& 	Offset	&	Yes	&	Yes &	Yes\\
& 	Apex	&	No	&	Yes &	Yes\\
\hline
\noalign{\smallskip}
\multicolumn{2}{l}{Number of coder}
&	2 & 2 & 3 \\
\hline
\noalign{\smallskip}
\multicolumn{2}{l}{Inter-coder reliability}
&	N/A & 0.846 & 0.82 \\
\hline

\end{tabular}
\end{center}
\end{table}

\subsubsection{SMIC}
The Spontaneous Micro-expression (SMIC) dataset comprises 16 subjects with 164 video clip.
The camera used to capture the video was PixeLINK PL-B774U with a temporal resolution of 100$fps$.
The cropped images have an average spatial resolution of 170 $\times$ 140 pixels, and each video consists of 34 frames (viz., 0.34$s$).
The ground-truths are labeled by two annotators, which include the emotion state, the action unit, the onset, offset frame indices.
However, the apex frame information of each video is not provided. 
The videos include three classes: positive (51 videos), negative (70 videos) and surprise (43 videos).
A three-class baseline recognition accuracy is reported as 48.78\% by employing LBP-TOP as the feature descriptor and SVM with Leave-One-Subject-Out Cross-Validation (LOSOCV) protocol.

\subsubsection{CASME II}
The Chinese Academy of Sciences Micro-Expression (CASME II) consists of 255 videos, elicited from 26 participants. 
The videos are recorded using Point Gray GRAS-03K2C camera which has a frame rate of 200$fps$.
The average video length is 0.34$s$, equivalent to 68 frames.
Each video's emotion label is annotated by two coders, where the reliability is 0.846.
All the images are cropped to $170\times 140$ pixels.
The ground-truth information provided by the database include the emotion state, the action unit, the onset, apex and offset frame indices.
The videos are grouped into seven categories: others (99 videos), disgust (63 videos), happiness (32 videos), repression (27 videos), surprise (25 videos), sadness (7 videos) and fear (2 videos).
A 5-class recognition baseline result of 63.41\% is reported which the feature extractor utilized was LBP-TOP and the classifier was Support Vector Machine (SVM) with Leave-One-Video-Out Cross-Validation (LOVOCV) protocol.
To perform cross database evaluation in the experiment later, some of the videos are recategorized based on the emotion state.
This is to cope with the database (i.e., SMIC) that has few expressions.
As a result, three main emotion classes are standardized: positive, negative and surprise.
Negative class include repression and disgust expressions; happiness is regarded as positive class, while the videos with others expression are not considered in the experiment.

\subsubsection{SAMM}
The Spontaneous Actions and Micro-Movements (SAMM) dataset contains 159 spontaneous videos, elicited from 32 participants. 
The videos are recorded using Basler Ace acA2000-340km camera with a temporal resolution of 200$fps$.
The average number of frames of the micro-expression video sequences is 74 frames (viz., 0.37$s$).
This dataset provides the cropped face video sequence with a spatial resolution of 400 $\times$ 400 pixels.
In an attempt to standardize the image resolution so that it is equivalently behaved as the other two databases, all the images are resized to 170 $\times$  140 pixels.
Each video is assigned with its emotion label, action unit, frame indices of apex, onset and offset.
The reliability of the marked labels by 3 coders is 0.82.
This database composes of eight classes of expressions: anger (57 videos), happiness (26 videos), other (26 videos), surprise (15 videos), contempt (12 videos), disgust (9 videos), fear (8 videos) and sadness (6 videos).
A recognition accuracy of 80.06\% is achieved with LBP-TOP as the feature extractor and Random Forest as the classifier with LOSOCV protocol.
For the experiment purpose, the videos are reclassified, such that it consists of three main classes: negative (i.e., anger, contempt, disgust, fear and sadness), positive (happiness) and surprise.
Note that videos with other expression are neglected.

\subsection{Experiment Settings}

In the OFF-ApexNet, the input features (i.e., $u$ and $v$) are resized into $[\aleph\times \aleph]=[28 \times28]$.
After that, they are processed by the convolutional, pooling, fully connected layers and finally the output layer.
The parameter setting for each layer is tabulated in Table~\ref{table:CNNsetting}.
To reduce the overfitting phenomena, a dropout regularisation operation is applied after the two fully connected layers.
A ratio of 0.5 is set, so that it keeps 50\% of the original output.
The initial learning rate is set to 0.0001 and a set of epochs values (i.e., 1000, 2000, 3000, 4000 and 5000) are examined.

Next, in the softmax classification layer, a cross-database micro-expression recognition will be performed, which means  the videos from the three databases are combined in the experiment.
Therefore, the total number of video involved in the experiment is 441, which is made up from SMIC (164 videos), CASME II (145 videos) and SAMM (132 videos).
There are three main emotion classes: negative, positive and surprise.
Specifically, a LOSOCV protocol is employed to examine the robustness of the proposed framework.
The principle of LOSOCV protocol is to iteratively leave out the videos of a single subject or participant as the testing set, while the rest of the videos will be served as training set.
This procedure is repeated for $k$ times, where $k$ is the number of participants in the experiment. 
Finally, the recognition results for all the participants are averaged to indicate the final recognition accuracy.
It should be reminded that, the video of the same subject will not be appearing in both the training and testing sets simultaneously.
Thus, it is considered as a person-independent approach.

To deal with the imbalance class distribution (i.e., 249 negative videos, 109 positive videos and 83 surprise videos), an alternative recognition performance measurement is exploited, namely F-measure.
Concretely, F-measure is defined as:
\begin{equation}\label{eq:f-measure}
\text{F-measure} := 2 \times \frac{\text{Precision} \times \text{Recall}}{\text{Precision + Recall}}, 
\end{equation}
for
\begin{equation}\label{eq:recall}
\text{Recall} := \frac{\text{TP}}{\text{TP + FN}}, 
\end{equation}
and
\begin{equation}\label{eq:precision}
\text{Precision} := \frac{\text{TP}}{\text{TP + FP}},
\end{equation}

\noindent 
where TP, FN and FP are the true positive, false negative and false positive, respectively.


\begin{table*}[tb] 
\begin{center}
\caption{OFF-ApexNet configuration for two convolution layers, two pooling layers, two fully connected layers and an output layer
}
\label{table:CNNsetting}
\begin{tabular}{lccccc}
\noalign{\smallskip}
\hline
\noalign{\smallskip}
Layer
& Filter size
& Kernel size
& Stride
& Padding
& Output size \\
\hline
\noalign{\smallskip}
Conv 1
& 5 $\times$ 5 $\times$ 1 
& 6
& [1,1,1,1]
& Same
& 28 $\times$ 28 $\times$ 6 \\

\noalign{\smallskip}
Pool 1
& 2 $\times$ 2 
& -
& [1,2,2,1]
& Same
& 14 $\times$ 14 $\times$ 6 \\

\noalign{\smallskip}
Conv 2
& 5 $\times$ 5 $\times$ 6 
& 16
& [1,1,1,1]
& Same
& 14 $\times$ 14 $\times$ 16 \\

\noalign{\smallskip}
Pool 2
& 2 $\times$ 2 
& -
& [1,2,2,1]
& Same
& 7 $\times$ 7 $\times$ 16 \\

\noalign{\smallskip}
FC 1
& -
& -
& -
& -
& 1024 $\times$ 1 \\

\noalign{\smallskip}
FC 2
& -
& -
& -
& -
& 1024 $\times$ 1 \\

\noalign{\smallskip}
Output
& -
& -
& -
& -
& 3 $\times$ 1 \\
\hline

\end{tabular}
\end{center}
\end{table*}

%% file: 6_results.tex
\section{Results and Discussion}  
\label{sec:results}
\subsection{Recognition Performance}
To the best of our knowledge, this is the first attempt that evaluates the feature extractor on three micro-expression databases.
Table~\ref{table:3db} reports the micro-expression recognition performance in both accuracy and F-measure of OFF-ApexNet method with various epoch size.
Concisely, all the three databases (i.e., SMIC, CASME II and SAMM) are merged and treated as a single database.
Therefore, a LOSOCV classification will be applied for $k=$68 times, which the 16 times are from SMIC, 24 times from CASME II and 28 times from SAMM.
From Table~\ref{table:3db}, it is noticed that OFF-ApexNet approach achieves the highest accuracy of 74.60\% and F-measure of 71.04\% when the epoch value is set to 3000.

\setlength{\tabcolsep}{5pt}

 \begin{table}[h]
 \begin{center}
 \caption{Overall micro-expression recognition accuracy and F-measure evaluated on SMIC, CASME II and SAMM databases using the proposed method, OFF-ApexNet}
 \label{table:3db}
 \begin{tabular}{ccc}
 \noalign{\smallskip}
 \hline
\noalign{\smallskip}
 Epoch
 & Accuracy (\%)
 & F-measure\\
 \hline

\noalign{\smallskip}
 1000
 & 72.56
 & .6905\\

\noalign{\smallskip}
 2000
 & 73.47
 & .7027\\

\noalign{\smallskip}
 3000
 & \textbf{74.60}
 & \textbf{.7104}\\

\noalign{\smallskip}
 4000
 & 72.79
 & .6918\\

\noalign{\smallskip}
 5000
 & 73.70
 & .6998\\
 \hline

 \end{tabular}
 \end{center}
 \end{table}


 On the other hand, Table~\ref{table:compare} shows the comparison of the micro-expression recognition performances of the proposed method (i.e., OFF-ApexNet) with other state-of-the-art feature extraction methods when evaluated on SMIC, CASME II and SAMM databases individually.
 Particularly, the previous research works (i.e., methods \#1 to \#11) focus to conduct the training and testing videos on a single database separately.
 For methods \#1 to \#11, the number of the expression to be predicted are based on the suggested expression category from the original papers \citep{casme2, smic, samm}.
 Some of the number of videos for certain expressions are quite few (i.e., less than 10 samples), thus those videos are neglected in the experiments.
 Concisely, there are a total of three expressions (i.e., positive, negative and surprise) in SMIC, five expressions (i.e., disgust, happiness, repression, surprise and others) in CASME II and five expressions (i.e., anger, happiness, contempt, surprise and other) in SAMM.

 In Table~\ref{table:compare}, method \#1 (i.e., LBP-TOP) is commonly known as the baseline approach in this automated micro-expression recognition domain, the recognition results reported are obtained by reproducing the experiments for each database.
 Since the SAMM database is released very recently, methods \#2 to \#11 did not examine the methods on this database.
 It can be seen that method \#11 (i.e., Bi-WOOF) outperformed the feature descriptors \#1 to \#10.  
 As such, Bi-WOOF approach is adopted to compare with the proposed method OFF-ApexNet later.

 To establish a fair comparison on the effectiveness of the proposed method, two state-of-the-art approaches (i.e., LBP-TOP and Bi-WOOF approach) are selected and the experimental configurations are set to similar across the comparing methods.
More precisely, the videos from the three databases (i.e., SMIC, CASME II and SAMM) are recategorized into exclusively three expressions (i.e., positive, negative and surprise).
As a result, the recognition performance is presented as methods \#12 and \#13.
 Particularly, for the proposed OFF-ApexNet approach (i.e., \#14), the feature extraction process follows the procedure as described in Section~\ref{sec:algorithm}.
 Firstly, the OFF-ApexNet model is trained by three databases as a whole using a LOSOCV strategy, then tested on each database separately.
 It is observed that, among all the methods shown in Table~\ref{table:compare}, OFF-ApexNet method achieves the best recognition results across all the three databases.

 
\setlength{\tabcolsep}{5pt}

\begin{table*}[tb]

\begin{center}

\caption{Comparison of micro-expression recognition performance in terms of \textit{Acc} (Accuracy (\%)) and \textit{F-mea} (F-measure) on the  SMIC, CASME II and SAMM databases for the state-of-the-art feature extraction methods, and the proposed method}

\label{table:compare}

\begin{tabular}{llcccccccc}
\noalign{\smallskip}
\hline
\noalign{\smallskip}
&
\multirow{2}{*}{Methods}
& \multicolumn{2}{c}{SMIC} 
& \multicolumn{2}{c}{CASME II} 
& \multicolumn{2}{c}{SAMM} \\

\cline{3-8}
\noalign{\smallskip}
   &  
 & Acc 
  & F-mea
 & Acc 
  & F-mea
 & Acc 
  & F-mea \\

  \noalign{\smallskip}
 \hline
  \noalign{\smallskip}

  &  
  & \multicolumn{2}{c}{3 classes} 
  & \multicolumn{2}{c}{5 classes} 
  & \multicolumn{2}{c}{5 classes} \\

 \noalign{\smallskip}
 \hline
 \noalign{\smallskip}

 1	
 & \begin{tabular}{@{}l@{}}LBP-TOP \\ \citep{smic,casme2,samm}\end{tabular}
 & 45.73	
 & .4600	
 & 39.68 
 & .3589
 & 35.56
 & .1768 \\

\noalign{\smallskip}
 2	
 &	OSF \citep{liong2014optical}	
 & 31.98	
 & .4461
 & -
 & - 
 & -
 & - \\

\noalign{\smallskip}
 3	
&	OSW \citep{liong2014subtle}
 & 53.05
 & .5431
 & 41.70
 & .3820
 & -
 & - \\

\noalign{\smallskip}
 4	
 & LBP-SIP \citep{wang2014lbp}
 & 54.88
 & .5502
 & 43.32
 & .3976
 & -
 & - \\

\noalign{\smallskip}
 5	
 & MRW \citep{oh2015monogenic}
 & 34.15
 & .3451
 & 46.15
 & .4307
 & -
 & - \\

\noalign{\smallskip}
 6	
 & STLBP-IP \citep{huang2015facial}
 & 57.93
 & .5829
 & 59.51
 & .5679
 & -
 & - \\

\noalign{\smallskip}
 7	
 & FDM \citep{xu2017microexpression}
 & 54.88
 & .5380
 & 41.96
 & .2972
 & -
 & - \\

\noalign{\smallskip}
 8
 & \begin{tabular}{@{}l@{}}Sparse Sampling \\ \cite{le2017sparsity} \end{tabular}
 & 58.00
 & .6000
 & 49.00
 & .5100
 & -
 & - \\

\noalign{\smallskip}
 9
 & STCLQP \citep{huang2016spontaneous}
 & 64.02
 & .6381
 & 58.39
 & .5836
 & -
 & - \\

\noalign{\smallskip}
 10
& MDMO \citep{liu2016main}
& -
& -
& 44.25
& .4416
& -
& - \\

\noalign{\smallskip}
11
& Bi-WOOF \citep{liong2018less}
& 61.59
& .6110
& 57.89
& .6125
& -
& - \\

\noalign{\smallskip}
\hline
\noalign{\smallskip}

&  
& \multicolumn{6}{c}{3 classes}  \\

\noalign{\smallskip}
\hline
\noalign{\smallskip}

\noalign{\smallskip}
12
& LBP-TOP
& 38.41
& .3875
& 60.00
& .5222
& 59.09
& .3640 \\

\noalign{\smallskip}
13
& Bi-WOOF \citep{liong2018less}
& 61.59
& .6110
& 80.69
& .7902
& 58.33
& .3970 \\

\noalign{\smallskip}
14
& \textbf{OFF-ApexNet}
& \textbf{67.68}
& \textbf{.6709}
& \textbf{88.28}
& \textbf{.8697}
& \textbf{68.18}
& \textbf{.5423}\\
\hline

\end{tabular}

\end{center}

\end{table*}


\subsection{Analysis and Discussion}

In Table 4, it can be seen that the accuracy result in SMIC database is the lowest among the three databases, when utilizing OFF-ApexNet. 
It might because of the apex frames of each video are spotted using an automatic apex spotting system, instead of utilizing the ground-truths. 
Referring to~\cite{liong2015automatic}, the average of frame difference between the detected and ground-truth apex is 13 frames. 
Thus, extracting the features from imprecise apex frame could affect the classification performance. 
For SAMM database, the F-measure is only 0.5423. 
This is due to the imbalance emotion class distribution where the ratio distribution is summarized in Table~\ref{table:distribution}.
SAMM database has the most severe imbalance data issue, whereby there are only 10\% surprise videos and 20\% positive videos.
It is also observed that the although SMIC is having balanced data distribution, the recognition performance (i.e., accuracy and F-measure) exhibited is lower than CASME II.
This is possibly due to the prominent expressive frames in SMIC database are not being captured by the camera as it has a much lower frame rate (i.e. 100$fps$), compared to CASME II (200$fps$). 
In a consequence, it fails to spot the precise apex frame in such circumstances.

\setlength{\tabcolsep}{5pt}

 \begin{table}[tb]
 \begin{center}
 \caption{Emotion ratio distribution of the three databases}
 \label{table:distribution}
 \begin{tabular}{lccccc}
        \noalign{\smallskip}
        \cline{2-4} 
        \noalign{\smallskip}
         & SMIC	 & CASME II	&	SAMM	\\
        \noalign{\smallskip}
        \hline
        \noalign{\smallskip}

        Negative  
        &	4	& 6		&	7	\\

        Positive 
        &	3	& 2		&	2	\\
        
        Surprise 
        &	3	& 2		&	1	\\
        \hline
 \end{tabular}
 \end{center}
 \end{table}
 
To further analyze the three-class recognition performance, confusion matrices are computed and shown in Table~\ref{table:cf_all} to~\ref{table:cf_samm}.
Generally, confusion matrix is a typical measurement to illustrate the classification rate for each expression.
The confusion matrix in Table~\ref{table:cf_all} indicates the overall performance, which means all the three databases are treated as a single database for training and testing purposes. 
The other three confusion matrices (i.e., in Table~\ref{table:cf_smic} to~\ref{table:cf_samm}) are tested on each database independently.
It can be seen that the negative emotion can always exhibit the highest prediction rate compared to positive and surprise.
The main reason is that, the negative emotion is the dominant class across the three databases (refer to Table~\ref{table:distribution}).

\setlength{\tabcolsep}{5pt}

 \begin{table}[tb]
 \begin{center}
 \caption{Confusion matrices of OFF-ApexNet for the recognition task on all the databases}
 \label{table:cf_all}
 \begin{tabular}{lccccc}
 
        \noalign{\smallskip}
        \cline{2-4} 
        \noalign{\smallskip}
         & Negative	 & Positive	&	Surprise	\\
        \noalign{\smallskip}
        \hline
        \noalign{\smallskip}

        Negative  
        &\bf.84	 & .11	&	.05	\\

        Positive 
        & .35	 &\bf.58	&	.07\\
        
        Surprise 
        & .20	 & .11	&\bf.69	\\
        \hline
 \end{tabular}
 \end{center}
 \end{table}

\setlength{\tabcolsep}{5pt}

 \begin{table}[tb]
 \begin{center}
 \caption{Confusion matrices of OFF-ApexNet for the recognition task on SMIC database}
 \label{table:cf_smic}
 \begin{tabular}{lccccc}
        \noalign{\smallskip}
        \cline{2-4} 
        \noalign{\smallskip}
         & Negative	 & Positive	&	Surprise	\\
        \noalign{\smallskip}
        \hline
        \noalign{\smallskip}

        Negative  
        &\bf.76	 & .17	&	.07	\\

        Positive 
        & .25	 &\bf.65	&	.10	\\
        
        Surprise 
        & .28	 & .14	&\bf.58	\\
        \hline
 \end{tabular}
 \end{center}
 \end{table}

\setlength{\tabcolsep}{5pt}

 \begin{table}[tb]
 \begin{center}
 \caption{Confusion matrices of OFF-ApexNet for the recognition task on CASME II database}
 \label{table:cf_casme}
 \begin{tabular}{lccccc}
        \noalign{\smallskip}
        \cline{2-4} 
        \noalign{\smallskip}
         & Negative	 & Positive	&	Surprise	\\
        \noalign{\smallskip}
        \hline
        \noalign{\smallskip}

        Negative  
        &\bf.93	 & .07	&	0	\\

        Positive 
        & .31	 &\bf.66	&	.03\\
        
        Surprise 
        & 0	 & 0	&\bf 1	\\
        \hline
 \end{tabular}
 \end{center}
 \end{table}

\setlength{\tabcolsep}{5pt}

 \begin{table}[tb]
 \begin{center}
 \caption{Confusion matrices of OFF-ApexNet for the recognition task on SAMM database}
 \label{table:cf_samm}
 \begin{tabular}{lccccc}
        \noalign{\smallskip}
        \cline{2-4} 
        \noalign{\smallskip}
         & Negative	 & Positive	&	Surprise	\\
        \noalign{\smallskip}
        \hline
        \noalign{\smallskip}

        Negative  
        &\bf.81	 & .10	&	.09	\\

        Positive 
        & .58	 &\bf.35	&	.08\\
        
        Surprise 
        & .33	 & .20	&\bf.47	\\
        \hline
 \end{tabular}
 \end{center}
 \end{table}

On the other hand, instead of utilizing both the horizontal and vertical optical flow component as the input data for OFF-ApexNet approach, the performance results for the individual flow component are also evaluated.
A comparison of the choice of input features is tabulated in Table~\ref{table:uv}, with a variation of the epoch values.
Concretely, $u$ is simply taking account only the horizontal optical flow features, while $v$ considers the vertical optical flow features.
$u+v$ refers to the proposed OFF-ApexNet method which fuses $u$ and $v$ motion information as the input data.
It is observed that OFF-ApexNet approach exhibits consistent high performance results compared to both the $u$ and $v$ methods.

From all the recognition performance shown, it is believed that the demonstration of OFF-ApexNet executes satisfactory recognition performance on the three micro-expression databases.


\setlength{\tabcolsep}{5pt}

\begin{table*}[tb]
\begin{center}
\caption{Comparison of the micro-expression recognition accuracy and F-measure when the input data to the network are: $u$, the horizontal optical flow features; $v$, the vertical optical flow featuers and $u+v$, both horizontal and vertical optical flow.}
\label{table:uv}
\begin{tabular}{ccccccc}
\noalign{\smallskip}
\hline
\noalign{\smallskip}

\multirow{2}{*}{Epoch} 
& \multicolumn{2}{c}{$u$} 
& \multicolumn{2}{c}{$v$} 
& \multicolumn{2}{c}{$u$+$v$}\\
\cline{2-7}
\noalign{\smallskip}

& Accuracy (\%)
& F-measure
& Accuracy (\%)
& F-measure
& Accuracy (\%)
& F-measure\\
 \hline

\noalign{\smallskip}

1000
& 67.35
& 0.6224
& 66.89
& 0.6199
& 72.56
& 0.6905\\
 \hline

\noalign{\smallskip}

2000
& 68.03
& 0.6307
& 67.57
& 0.6253
& 73.47
& 0.7027\\
 \hline

\noalign{\smallskip}

3000
& 66.21
& 0.6134
& 65.31
& 0.6047
& \textbf{74.60}
& \textbf{0.7104}\\
 \hline

\noalign{\smallskip}

4000
& 67.35
& 0.6287
& 65.99
& 0.6111
& 72.79
& 0.6918\\
 \hline

\noalign{\smallskip}

5000
& 66.67
& 0.6220
& 66.21
& 0.6150
& 73.70
& 0.6998\\
\hline

\end{tabular}
\end{center}
\end{table*}


%% file: 7_conclusion.tex
\section{Conclusion} 
\label{sec:conclusion}

In a nutshell, a novel feature extraction approach, Optical Flow Features from Apex frame Network (OFF-ApexNet) is introduced to recognize the micro-expressions.
As its name implies, it combines both the handcrafted features (i.e., optical flow derived components) and the fully data-driven architecture (i.e., convolutional neural network).
First, the horizontal and vertical optical flow features are computed from onset and apex frames.
Then, the features are proceed to feed into a neural network to further highlight significant expression information.
The utilization of both the handcrafted and data-driven features is capable to achieve promising performance results on three recent state-of-the-art databases, namely SMIC, CASME II and SAMM.
Note that this is the first attempt for cross-dataset validation on three databases in this domain.
As a result, a highest three-class classification accuracy of 74.60\% was achieved with its F-measure of 0.71, when considering the three databases as a whole.

The contributions of this work point to some avenues for further research.
For instance, rather than utilizing optical flow feature, other feature extractors (i.e., LBP, HOG, SIFT, etc.) can be applied to better represent the motion details.
As a result, valuable input data will be passed to the convolutional neural network architecture for feature enrichment and selection, thereby improve the classification performance.
Besides, attention can be devoted to handling the issues of imbalance data in these databases so that the methods proposed can lead to consistent good recognition results across all the expressions.